# WindowSHAP: An Efficient Framework for Explaining Time-series Classifiers based on Shapley Values

*Amin Nayebi[1,\*], Sindhu Tipirneni[2], Chandan K Reddy[2], Brandon Foreman[3], Vignesh Subbian[1]*

[1] Department of Systems and Industrial Engineering, University of Arizona, AZ, USA

[2] Department of Computer Science, Virginia Tech, VA, USA

[3] College of Medicine, University of Cincinnati, OH, USA

**Abstract**

Unpacking and comprehending how black-box machine learning algorithms (such as deep learning models) make decisions has been a persistent challenge for researchers and end-users. Explaining time-series predictive models is useful for clinical applications with high stakes to understand the behavior of prediction models, e.g., to determine how different variables and time points influence the clinical outcome. However, existing approaches to explain such models are frequently unique to architectures and data where the features do not have a time-varying component. In this paper, we introduce *WindowSHAP*, a model-agnostic framework for explaining time-series classifiers using Shapley values. We intend for *WindowSHAP* to mitigate the computational complexity of calculating Shapley values for long time-series data as well as improve the quality of explanations. *WindowSHAP* is based on partitioning a sequence into time windows. Under this framework, we present three distinct algorithms of *Stationary*, *Sliding* and *Dynamic WindowSHAP*, each evaluated against baseline approaches, KernelSHAP and TimeSHAP, using perturbation and sequence analyses metrics. We applied our framework to clinical time-series data from both a specialized clinical domain (Traumatic Brain Injury - TBI) as well as a broad clinical domain (critical care medicine). The experimental results demonstrate that, based on the two quantitative metrics, our framework is superior at explaining clinical time-series classifiers, while also reducing the complexity of computations. We show that for time-series data with 120 time steps (hours), merging 10 adjacent time points can reduce the CPU time of *WindowSHAP* by 80% compared to KernelSHAP. We also show that our *Dynamic WindowSHAP* algorithm focuses more on the most important time steps and provides more understandable explanations. As a result, *WindowSHAP* not only accelerates the calculation of Shapley values for time-series data, but also delivers more understandable explanations with higher quality.

# 1   Introduction

Explaining and understanding the decision-making process of black-box machine learning algorithms is one of the major challenges for the research community in computing and information sciences. Despite the strong performance of these algorithmic predictions, their non-linear structure makes them challenging to discern what information in the input data causes them to generate particular predictions that support clinical decisions [1]. Rationalizing model behavior can help uncover biases, promote fairness and transparency, and most importantly, increase trust among end-users [2], [3]. Furthermore, modern privacy laws such as the European Union General Data Protection Regulation (GDPR) emphasizes users' right to explanations related to automated decision-making [4]. These trends and challenges, collectively, necessitate the development of tools for elucidating black-box clinical prediction models.

Model explainability is further complicated by the emerging shift in underlying data from using relatively simple, abstracted clinical data to more complex, routinely collected, longitudinal clinical data. During the course of a patient's care process, electronic health records (EHRs) host vast amounts of time-series data through frequent charting of vital signs, laboratory tests, and prescriptions [5]. Such time-series data has the potential to support clinical decision-making and forecast a variety of patient outcomes such as clinical deterioration, functional improvement, discharge disposition, and effectiveness of interventions[6]–[9].

Existing explainability methods are often focused on extracting the importance or contribution of input features to the model prediction. For example, SHAP (SHapley Additive exPlanations) [10] is one such method that generates contribution scores (i.e., Shapley values) to explain the individual predictions based on a coalitional game theory that satisfies three desirable properties for explanations: consistency, local accuracy, and missingness. There exists an alternative, kernel-based approximation of Shapley values called KernelSHAP that can reduce the complexity of calculating Shapley values by sampling from a smaller number of feature subsets [10].

Despite its prominence in informatics and clinical applications [11]–[16] , SHAP is not entirely appropriate for time-series predictive models. First, it was not originally intended to be used with time-series data. Second, while KernelSHAP provides a model-agnostic approximation of Shapley values that sets a ceiling for the number of sampled feature subsets, it is still computationally expensive for high-dimensional data [17], [18]. Last but not least, sequential data points in clinical time-series data are often highly dependent on each other, which can lead to misleading KernelSHAP results [19]. When there are several highly dependent features (e.g., variable-time point pairs in time-series data), the joint contribution of these features is distributed among them, resulting in a large number of small Shapley values [18]. This makes it more difficult to visualize data or extract useful explanations from the contribution scores.

The primary objective of this research is to design and evaluate an explanation method based on Shapley values that is (1) applicable to time-series data, (2) computationally feasible for high-resolution time-series data, and (3) able to tackle dependencies between sequential data points. To address the shortcomings of KernelSHAP, we present *WindowSHAP*, a framework designed to explain time-series prediction models more effectively and accurately. *WindowSHAP* reduces the total number of features for which Shapley values must be determined by combining neighboring time steps into a time window. Instead of calculating Shapley values for every possible time step and variable combinations, we simply calculate Shapley values for each time window (see **Figure 1** for conceptual demonstration). We propose various types of time windows, each with their own advantages under the *WindowSHAP* framework. For evaluation purposes, we train three deep learning models on time-series data both from a specialized clinical domain (Traumatic Brain Injury) and a broad clinical domain (critical care medicine) to show the applicability of algorithms. We compare our proposed framework with competitive baselines using different quantitative metrics to demonstrate the efficiency of our algorithms and accuracy of their explanations.

In summary, the main contributions of this study are as follows:

- Developing the *WindowSHAP* framework, a variation of Shapley additive explanations for time-series data.
- Introducing and evaluating variations of *WindowSHAP* based on different windowing techniques in both categories of fixed- and variable-length time windows.
- Validating our method on real-world clinical time-series data by employing a variety of quantitative metrics.

The rest of the paper is organized as follows. Section 2 describes the related research. Section 3 presents the *WindowSHAP* framework as well as the datasets and prediction models that are used in experiments. Section 4 details our results, Section 5 discusses the findings and implications of our work, and Section 6 concludes our work.

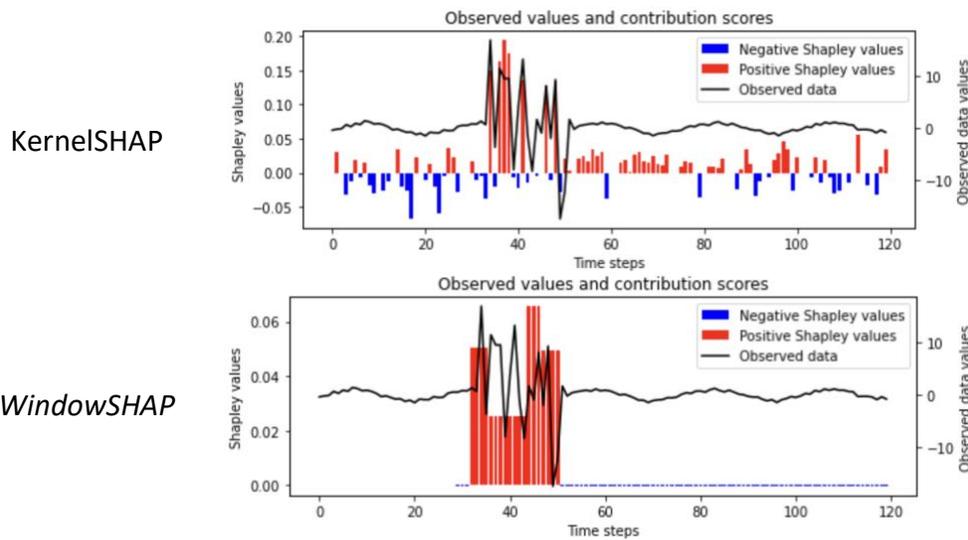

*Figure 1.* Conceptual Demonstration of KernelSHAP vs WindowSHAP for a classification model for an individual instance, predicting whether there is an anomaly in a synthetically generated sequence. The top picture shows the sequence and its Shapley values derived from the KernelSHAP while the bottom depicts the Shapley values from WindowSHAP. While KernelSHAP is spreading the Shapley values all over the sequence, our approach focuses more on the part of the sequence that is more important, avoiding calculating Shapley values for each single time step.

## 2  Related Work

Explanation approaches can be broadly classified into model-specific and model-agnostic techniques. Model-specific techniques generate explanations for model behavior using architectural properties of the model. Attention mechanism is the most commonly utilized model-specific strategy for explaining deep learning time-series classifiers [20]–[24], where an importance score is assigned to each time step using attention layers. Gradients can also be used to describe deep learning time-series classifiers by allocating a weight to each input based on the outcome's derivative with respect to the corresponding feature [25]. One study [26] produced important ratings for deep learning models using "difference from reference" rather than gradients. The reference value represents a default or neutral input, and the algorithm back-propagates the "difference from reference" values to avoid artifacts such as gradient discontinuity.

Model-agnostic explanation methods can be used on any black-box model regardless of the model structure. They only rely on the inputs and outputs rather than the model's internal architecture. In recent years, one of the most common model-agnostic approaches is to attribute importance to features using Shapley values [10]. The Shapley value is a concept from coalitional game theory that fairly distributes the payout generated by a game to each of its players [27]. To calculate the Shapley value for a single feature, $2^{D-1}$ terms must be calculated. Hence, when the total number of input features increases, as it does in modern machine learning applications, the complexity of calculating Shapley values grows exponentially. Some approximations for calculating Shapley values have been developed, including KernelSHAP, which sums over a smaller sample of feature subsets rather than all subsets [10]. However, as the number of features increases, so does the number of sampled subsets (exponentially) in order to maintain an adequate accuracy for the approximated Shapley values.

TimeSHAP [28] is a temporal model extension of SHAP that includes a pruning mechanism that combines all initial time steps whose cumulative relevance is less than a certain threshold. It assumes that the earliest time steps in sequential data sets are the least important time points. However, this assumption can be criticized because it might not be true in all cases of time-series data. For example, the conceptual demonstration in **Figure 1** shows a situation of anomaly detection where the most important time points are not necessarily at the end of a sequence.

Temporal Importance Model Explanation (TIME) [29] was developed to identify critical temporal steps and time intervals at the global explanation level. This approach, however, is confined to providing generalized, global significance of time steps and cannot be used for a single instance of data. Even though all of these methods are applicable to time-series classifiers, they either do not provide local explanations for a single instance of data, or they do not handle high dimensionality of time-series as well as high dependency of adjacent time steps.

## 3 Methods

### 3.1 Shapley values for tabular data

Shapley values assign an importance (contribution) score $\phi_i$ to the $i^{th}$ feature, indicating how much the model output for a single instance is influenced by its $i^{th}$ feature. Based on [10], we provide a formulation of Shapley values. Assuming that $x \in R^D$ is an input of a prediction model $f(\cdot)$, the Shapley value for feature $i$ for a given input $x = x^*$ is calculated by

$$\phi_i = \sum_{S \subset \Delta \setminus \{i\}} \frac{|S|!\,(D - |S| - 1)!}{D!} [v_{x^*}(S \cup \{i\}) - v_{x^*}(S)] \quad (1)$$

where $\Delta$ is the set of all features, $S$ is a subset of feature indices, and $v_{x^*}(S)$ is the characteristic function which shows the output of the prediction function if only features in set $S$ are present from input $x^*$. The characteristic function is defined as follows:

$$v_{x^*}(S) = E[f(x)|x_S = x_S^*] \quad (2)$$

Here, $x_S$ is a sub-vector of $x$ representing the features in set $S$. Due to the local accuracy property of Shapley values, the sum of all feature importance scores is equal to the prediction model output, i.e., $f(x^*) = \phi_0 + \sum_{i=1}^{D} \phi_i$ where $\phi_0$ is the output of characteristic function when all the features are absent. See **Table 1** for a description of the notations used in this work.

*Table 1. List of notations used in the paper*

| Notation | Description |
|---|---|
| $f(\cdot)$ | The prediction model |
| $x$ | Input of the prediction model |
| $S$ | A subset of all features |
| $\Delta$ | The set of all combinations of variables and time steps |
| $\Delta^i$ | The set of all time steps for variable $i$ |
| $\omega_k^i$ | $k^{th}$ time window in the variable $i$'s sequence |
| $w_i$ | The number of time windows that variable $i$ is partitioned to |
| $\Omega^i$ | The set of all windows for variable $i$ |
| $\Omega$ | The set of all time windows for all variables |
| $\phi_i$ | The contribution score of variable $i$ |
| $\phi_{(i,t)}$ | The contribution score of variable $i$ at time point $t$ |
| $\phi_{\omega_k^i}$ | The contribution score assigned to the $k$th window in variable $i$ |
| $l$ | Window length parameter in *Stationary* and *Sliding WindowSHAP* algorithms |
| $s$ | Stride parameter in *Sliding WindowSHAP* algorithm |
| $\delta$ | The Shapley value threshold in *Dynamic WindowSHAP* algorithm |
| $n_w$ | Maximum number of time windows in *Dynamic WindowSHAP* algorithm |

## 3.2 Shapley values for time-series data

The general Shapley values formulation provided in equation (*1*) is not directly applicable to time-series data. In order to calculate Shapley values for time-series data, each possible combination of variable and time step is considered an input feature, which results in Shapley values for each of these combinations. Suppose that $X \in R^{D \times L}$ is a time-series instance with $D$ variables and $L$ time steps. Defining $\Delta = \{(i,t) : 1 \leq i \leq D, \ 1 \leq t \leq T\}\}$ as the set of all combinations of variables and time steps, we calculate the Shapley value of variable $i$ at time point $t$ as

$$\phi_{(i,t)} = \sum_{S \subset \Delta \setminus \{(i,t)\}} \frac{|S|! \, (D \times L - |S| - 1)!}{(D \times L)!} [v_{X^*}(S \cup \{(i,t)\}) - v_{X^*}(S)] \qquad (3)$$

where $v_{X^*}(S)$ is the characteristic function which denotes the prediction output when only the variable-time pairs in set $S$ are present in input $X^*$. Extracting Shapley values for high resolution time-series will be very time consuming since for each pair of $(i,t)$, $2^{D \times L - 1}$ terms should be calculated.

## 3.3 WindowSHAP

We introduce our efficient framework called *WindowSHAP* to estimate Shapley values for time-series data in this section. *WindowSHAP* is designed on the idea of constructing windows from either nearby or non-adjacent temporal steps. In this method, we compute Shapley values for each individual time window rather than for all possible combinations of variable-time points. Assume that we partition $\Delta^i = \{(j,t) \in \Delta : j = i\}$ into $w_i$ non-overlapping time windows. Note that a window need not necessarily have a contiguous set of time points. The resulting set of windows for variable $i$ is represented as $\Omega^i = \{\omega_1^i, \omega_2^i, \ldots, \omega_{w_i}^i\}$ where $\omega_k^i \subset \Delta^i$ shows the $k^{th}$ time window in the variable $i$'s sequence. Considering each window for each variable as a feature, the Shapley value for the $k^{th}$ time window of variable $i$ is calculated as

$$\phi_{\omega_k^i} = \sum_{S \subset \Omega \setminus \omega_k^i} \frac{|S|! \, (|\Omega| - |S| - 1)!}{|\Omega|!} [v_{X^*}(S \cup \omega_k^i) - v_{X^*}(S)] \qquad (4)$$

where $\boldsymbol{\Omega}$ is the set of all time windows for all variables i.e., $\boldsymbol{\Omega} = \bigcup_{i=1}^{D} \boldsymbol{\Omega}^i$. The Shapley value of any variable-time point combination can be estimated by distributing the importance of a time window equally among its time points, i.e.,

$$\phi_{(i,t)} = \frac{\phi_{\omega_k^i}}{|\omega_k^i|}, \quad \forall (i,t) \in \omega_k^i \tag{5}$$

The Shapley value of all windows in $\boldsymbol{\Omega}$ add up to the prediction model output based on the local accuracy property. The local accuracy property is maintained after dispersing the Shapley values of time windows among their time points, i.e.,

$$\phi_0 + \sum_{\omega_k^i \in \boldsymbol{\Omega}} \sum_{(i,t) \in \omega_k^i} \phi_{(i,t)} = \phi_0 + \sum_{\omega_k^i \in \boldsymbol{\Omega}} \phi_{\omega_k^i} = f(\boldsymbol{x}^*) \tag{6}$$

Under the *WindowSHAP* framework, we describe three algorithms: (1) *Stationary WindowSHAP*, (2) *Sliding WindowSHAP*, and (3) *Dynamic WindowSHAP*. *Stationary WindowSHAP* and *Sliding WindowSHAP* are our fixed-length algorithms where all time windows are of the same length, while *Dynamic WindowSHAP* is a variable-length algorithm. We will describe each algorithm in the following sections.

### 3.3.1 Stationary WindowSHAP

In this approach, the time-axis is segmented into fixed-length windows. Even though all time windows have the same length, if the length of the sequence is not divisible by the length of the time window, the last time window may be smaller than the others. **Figure 2** shows a partitioning the time-axis for the *Stationary WindowSHAP* algorithm.

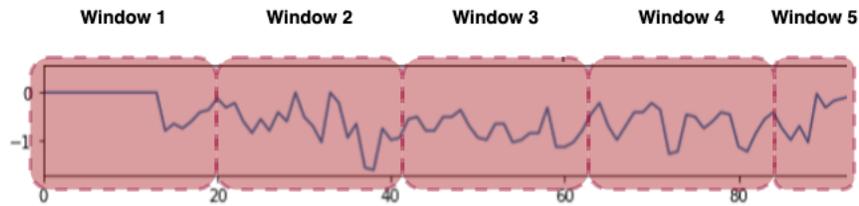

*Figure 2. A visualization of time-axis partitioning in the Stationary WindowSHAP algorithm. The windows are non-overlapping, contiguous, and of the same length, except possibly the last window being smaller.*

---

**Algorithm 1** *Stationary WindowSHAP*

**Input:** Input sequence $X$, prediction model $f$, window size $l$

**Output**: Shapley values $\Phi$

| | |
|---|---|
| $w_i \leftarrow \left\lceil \frac{L}{l} \right\rceil, \quad \forall i = 1, \dots, D$ | ❖ Calculating the total number of time windows |
| $\omega_k^i \leftarrow \{(i,t) \mid (k-1) \cdot l < t \leq \min\{L, k \cdot l\}\}$ $\forall k = 1, \dots, w_i, \forall i = 1, \dots, D$ | ❖ Building the time windows using adjacent time steps |
| $\Omega \leftarrow \cup \, \omega_k^i$ | ❖ Building the set of all time windows |
| $\Phi \leftarrow WindowSHAP(model = f, input = X, Windows = \Omega)$ | ❖ Calculating the Shapley values for all time windows using *WindowSHAP* |
| return $\Phi$ | |

### 3.3.2 Sliding WindowSHAP

Since the stationary windowing approach may not explain time points near the boundary of neighboring windows, we developed sliding time window approach (see Algorithm 2) where adjacent time windows overlap. Since in the WindowSHAP framework, $\Omega^i$ should have non-overlapping time windows for variable $i$, we shift the time window to the end of the sequence over iterations. This algorithm iteratively divides the temporal sequence into inside and outside of the specified time window, resulting in two Shapley values for each sequence. The algorithm's window length ($l$) and stride ($s$) parameters, respectively, determine the length and the amount of shift for time windows in each iteration.

**Algorithm 2** *Sliding WindowSHAP*

**Input:** Input sequence $X$, prediction model $f$, window size $l$, stride $s$
**Output**: Shapley values $\Phi$

| | |
|---|---|
| $n_w \leftarrow \left\lceil \frac{L-l}{s} \right\rceil + 1$ | ❖ Calculating the total number of sliding time windows |
| $\Phi \in \mathbb{R}^{D \times n_w}$ | ❖ Initializing Shapley values for all possible time windows and features |
| **for** $j \in \{0, 1, 2, \dots, n_w - 1\}$ **do** | ❖ Iterating over all time windows |
| $\quad \omega_1^i \leftarrow \{(i,t) \mid j \cdot s + 1 \leq t \leq j \cdot s + l - 1\}$ $\quad (\forall i = 1, \dots, D)$ | ❖ Set of time steps inside the time window |
| $\quad \omega_2^i \leftarrow \Delta^i - \omega_1^i$ $\quad (\forall i = 1, \dots, D)$ | ❖ Set of time steps outside the time window |
| $\quad \Omega \leftarrow \cup \, \omega_k^i$ | ❖ Building the set of all time windows |

$$W \leftarrow WindowSHAP(\\
\quad model = f,\\
\quad input = X,\\
\quad Windows = \Omega)$$

❖ Calculating the Shapley values for all features inside and outside the time window

$$\Phi_{:,j+1} \leftarrow W_{:,1}$$

❖ Updating the matrix of Shapley values for the corresponding time window

return $\Phi$

---

**Figure 3** demonstrates how *Sliding WindowSHAP* works (a) in each iteration and (b) after all iterations are completed. After all iterations are completed, the Shapley value of each time point is computed by averaging the Shapley values of the time windows that contain the time point. For example, the Shapley value of time step $t$ in **Figure 3-b** is $\frac{\phi_2+\phi_3}{2l}$.

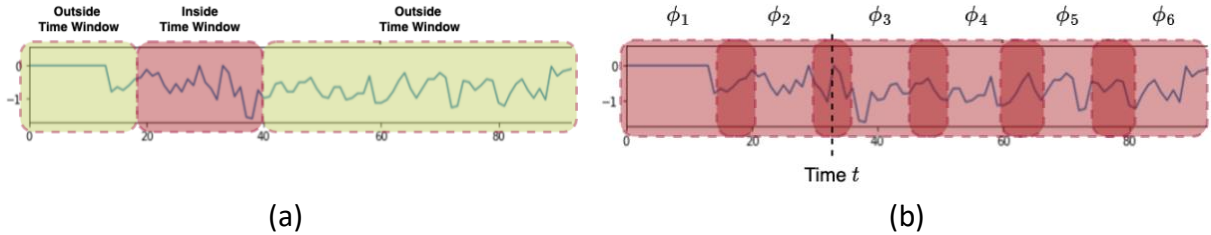

(a)            (b)

*Figure 3. Demonstration of Sliding WindowSHAP: (a) depicts a single iteration in which the entire sequence is divided into two time intervals, inside and outside of the time window. (b) shows the final windowing result after all iterations have been completed and a Shapley value has been produced for each time window.*

### 3.3.3 Dynamic WindowSHAP

In this approach, we divide the entire series into variable-length time windows. To accomplish this, we first define what the optimal split is using the following two objectives:

1. Keeping the number of time windows as few as possible to avoid increasing the algorithm's complexity
2. Avoiding lengthy windows with large contribution scores to minimize information loss

The Shapley value of all time windows is calculated in each iteration of this algorithm, and time windows with Shapley values greater than a threshold, $\delta$, are split into two subsequences. The method terminates when either it reaches the maximum number of time windows, $n_w$, or the Shapely values of all time windows are less than or equal to $\delta$. In the initial iteration of the algorithm, each time-series is considered as a single time window. For simplicity, Algorithm 3 shows the pseudo code of this method for a univariate sequence ($x \epsilon R^L$). However, this can be easily extended to include all features of a multivariate time-series data at once.

**Algorithm 3** *Dynamic WindowSHAP*

**Input:** Input sequence $x$, prediction model $f$, Shapley value threshold $\delta$, maximum number of time windows $n_w$
**Output**: Shapley values $\Phi$

| | |
|---|---|
| $S_{new} \leftarrow \{1, L\}$, $S \leftarrow \{\}$ | ❖ Initializing the set $S$ which stores splitting points of the sequence |
| **While** $S_{new} \neq S$ and $\|S_{new}\| \leq n_w+1$ | ❖ Stop the iterations when stopping criteria are met |
| $\quad S \leftarrow S_{new}$ | |
| $\quad \omega_k^i \leftarrow \{(i,t) \| S_k \leq t \leq S_{k+1}\}$ $(\forall k = 1, \ldots, \|S\| - 1)$ | ❖ Building the time windows based on the split points |
| $\quad \Omega \leftarrow \bigcup \omega_k^i$ | ❖ Building the set of all time windows |
| $\quad \Phi \leftarrow WindowSHAP($ $\quad\quad model = f,$ $\quad\quad input = X,$ $\quad\quad Windows = \Omega)$ | ❖ Calculating Shapley values for all time windows using *WindowSHAP* |
| $\quad$ **for** $k \in \{1, 2, \ldots, \|S\|\}$ **do** | |
| $\quad\quad$ **if** $\Phi_k > \delta$ **do** | ❖ Adding a new split point to $S$ if an interval's Shapley value is larger than the threshold |
| $\quad\quad\quad new\_point \leftarrow \lfloor \frac{S_{k-1}+S_k}{2} \rfloor$ | |
| $\quad\quad\quad$ **add** $new\_point$ **to** $S_{new}$ | |
| **return** $\Phi$ | |

**Figure 4** demonstrates how the algorithm works in four iterations of an example where, in the second iteration, only the second time window has a Shapley value greater than $\delta$. Hence in the next iteration, this window gets split into two equal time windows. The algorithm terminates at iteration four because all Shapley values for time windows are less than $\delta$.

### 3.4 Evaluation metrics

Given the volume and variety of clinical time-series data, evaluating and confirming these explanations through direct inspection by domain experts is not practicable. To implement a fair and quantitative evaluation of the explanation results, we adopt metrics discussed by Schlegel et al.[30]. They propose two metrics - perturbation and sequence analysis metrics - for evaluating explanations of single time points and temporal patterns, respectively. These metrics are defined based on the assumption that if a relevant/important feature (at a certain time point) changes, the performance of an accurate prediction model must decrease.

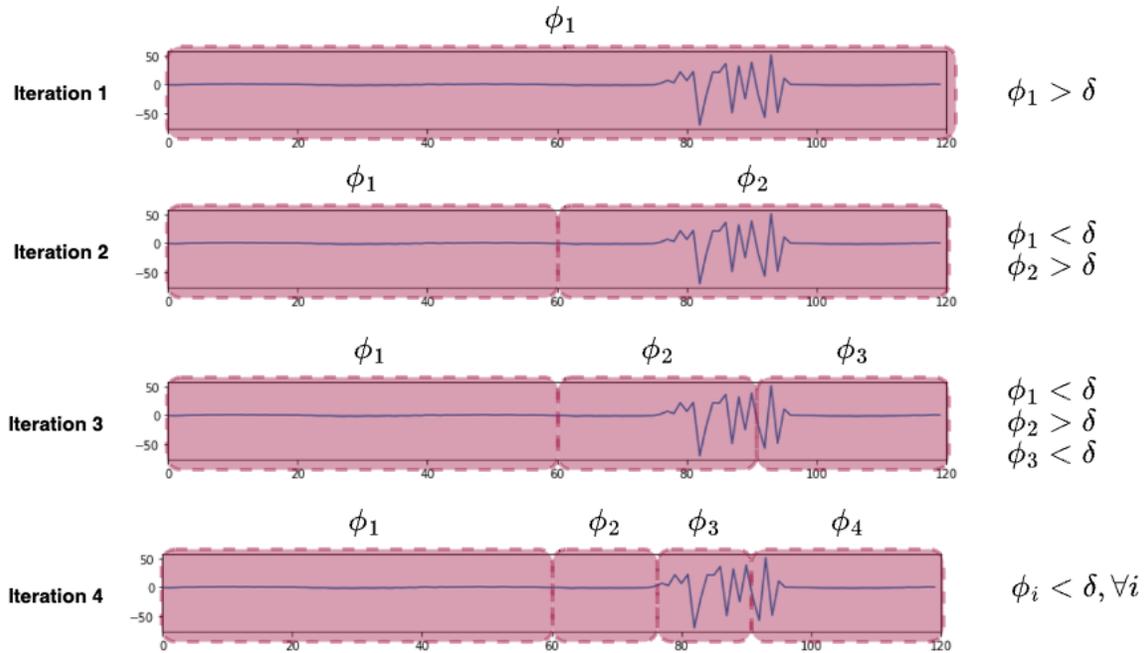

**Figure 4.** Demonstration of Dynamic WindowSHAP algorithm for a sequence. The algorithm stops in the fourth iteration because all the Shapley values for time windows are less than the threshold $\delta$

**Perturbation analysis metric:** For a univariate time-series $t = (t_1, t_2, \ldots, t_T)$ and the relevance vector $r = (r_1, r_2, \ldots, r_T)$, a time point $t_i$ changes to $(\max(t) - t_i)$ if the corresponding $r_i$ is larger than the $p^{th}$ percentile of $r$. The new sequence is called $t^{inverse}$ for which the model quality metric ($qm$) is calculated, e.g., loss function. The difference between quality metric for the original sequence and the perturbed sequence is called the *perturbation analysis metric*. We can calculate this as a percentage of change in the quality metric, i.e., $\frac{qm(t^{inverse})}{qm(t)}$. Perturbation analysis metric aims to evaluate the impact of perturbing individual time points on model performance. However, this approach does not directly consider the impact of temporal patterns or trends, such as slopes or minima, on model performance.

**Sequence analysis metric:** Unlike the perturbation analysis metric, this metric focuses on the inter-dependency of time points in a sequence and how model performance is affected when entire segments of the time-series are replaced or modified. Sequence analysis metric helps assess the ability of the model and the explanation method to capture and explain the significance of temporal patterns in the data.

For a univariate time-series $t = (t_1, t_2, \ldots, t_T)$ and the relevance vector $r = (r_1, r_2, \ldots, r_T)$, a time point $t_i$ in the sequence is chosen if the corresponding $r_i$ is larger than the $p^{th}$ percentile of $r$. Then, the time interval $(t_i, t_{i+1}, \ldots, t_{i+n})$ is replaced with the mean of the sequence and the resulting time-series is called $t^{mean}$. Similar to perturbation analysis, the difference between quality metric for the original sequence and the new sequence is called the *sequence analysis metric*. We can calculate this as a percentage of change in the quality metric, i.e., $\frac{qm(t^{mean})}{qm(t)}$.

In summary, the perturbation analysis metric evaluates the influence of individual time points on the model's performance, while the sequence analysis metric assesses the impact of temporal patterns and trends. Both metrics together provide a more comprehensive evaluation of explanaton methods, ensuring that they account for both individual time points and temporal patterns in time series data.

### 3.5  Data sources

To test the model-agnostic explanation methods (e.g., *WindowSHAP*), we used three distinct clinical time-series data sets to develop and train three different deep learning prediction models. Two sets of clinical time-series data were derived from the prospective, multicenter Transforming Research and Clinical Knowledge in Traumatic Brain Injury (TRACK-TBI) study [31], while the third dataset came from MIMIC III EHR data [32]. TRACK-TBI includes detailed clinical data on nearly 3,000 Traumatic Brain Injury (TBI) patients from 18 academic Level I trauma hospitals throughout the United States. MIMIC III is a de-identified EHR data for nearly 40,000 intensive care unit (ICU) patients at Beth Israel Deaconess Medical Center, Boston, MA. We utilized both specialized and generic clinical domain data to demonstrate the applicability of our method to a wide range of clinical areas.

We used two subsets of data from the TRACK-TBI study to develop two distinct prediction models: (1) time-series EHR data collected during hospital stay and (2) high-resolution physiologic data. EHR time-series data are comprised of clinical variables collected during the initial five days of hospital stay across patients admitted to the hospital with TBI. We included 900 out of 2996 participants, who had outcome data and recordings of blood pressure for at least 12 hours in the first 48 hours of ICU stay. We developed a prediction model to predict the long-term functional outcome of patients using the Glasgow Outcome Scale-Extended, a categorical outcome measure ranging from 1 (death) to 8 (upper good recovery) and dichotomous as good outcome (GOSE 5-8) vs poor outcome (GOSE 1-4). Detailed information on this dataset and the prediction model can be found in [33].

A subset of participants in the TRACK-TBI study (n = 25) also had high-resolution recordings of physiologic data using a bedside data aggregation system (Moberg Solutions, Inc; Ambler, PA). The waveform data for these individuals includes vital signs such as heart rate and arterial blood pressure, as well as intracranial monitoring data. A prediction model was developed and trained to predict an adverse event. Here, an adverse event is defined when intracranial pressure (ICP) is larger than 22 mmHg for at least 15 minutes.

The third clinical prediction model is based on the MIMIC data set and uses the initial 48 hours of clinical data to predict patient mortality in the subsequent 48 hours. The MIMIC time-series data includes eight vital signs and twenty lab measurements. The missing values in vital signs were imputed using the mean, whereas forward imputation is employed for missing laboratory measurements, i.e., the lab values were retained until a new measurement is obtained. A summary of data and model characteristics are included in **Table 2**.

*Table 2. Datasets characteristics*

| Dataset characteristic | TRACK-TBI EHR dataset | TRACK-TBI physiologic dataset | MIMIC-III dataset |
|---|---|---|---|
| Size of data (#samples, #time steps, #variables) | (900, 120, 62) | (5,816, 360, 8) | (22,988, 48, 26) |
| Duration of each time step | 1 hour | 10 seconds | 1 hour |
| Types of features | Vitals, lab measurements, GCS score components | Vital signs and intracranial data | Vital signs and lab measurements |
| Outcome | Dichotomized GOSE score after 6 months | Adverse event of high ICP values (binary outcome) | Mortality after 48 hours |
| Unfavorable Label (%) | 22% | 8.8% | 10% |

| Method of handling missing values | Imputation during training using GRU-D units | Imputation using linear interpolation | Forward imputation |

To evaluate the quality of explanations generated by our algorithms, we compared the results to those of KernelSHAP and TimeSHAP, two baselines. We utilized all three RNN-based prediction models constructed and trained on distinct clinical time-series datasets, including TBI EHR data, TBI physiologic data, and MIMIC-III data. For each dataset, 50 random samples were selected from the test dataset and explanations were generated using several techniques. We computed perturbation analysis and sequence analysis scores for each combination of prediction model and explanation algorithm.

### 3.6 Implementation details

We developed three Recurrent Neural Network (RNN)-based prediction models and trained on distinct clinical time-series datasets, including TBI EHR data, TBI physiologic data, and MIMIC-III data. All algorithms and prediction models were implemented in Python 3 environment and is available online [34]. The prediction models were developed using Keras library. The detailed specifications of each prediction model are described in **Table 3**.

*Table 3.* Prediction models specifications

| RNN model characteristic | Data sets | | |
| --- | --- | --- | --- |
| | **TBI EHR data** | **TBI physiologic data** | **MIMIC-III data** |
| Number of RNN layers | 100 (GRU-D units) | 200 (GRU units) | 70 (GRU units) |
| Number of neurons in the (first, second) layer after RNN | (50, 0) | (70, 30) | (40, 10) |
| Loss function optimization algorithm | Adam | Adam | Adam |
| Learning rate | 0.0002 | 0.0002 | 0.0002 |
| Regularization rate | 0.208 | 0.004 | 0.01 |
| RNN dropout rate | 0.42 | 0.3 | 0.4 |
| RNN recurrent dropout rate | 0.58 | 0.3 | 0.4 |
| Hidden layer dropout rate | 0.29 | 0.3 | 0.4 |
| Batch size | 32 | 32 | 64 |

To evaluate the quality of explanations generated by our algorithms, we compared the results to those of KernelSHAP and TimeSHAP, two baselines. To extract the Shapley values for each variable-time step combination, we modified the implementation of TimeSHAP such that the Shapley values of pruned variables and time steps are distributed uniformly among them. We utilized a grid search to determine the optimal parameter values for all explanation methods (see **Table 4**). For each dataset, 50 random samples were selected from the test dataset and explanations were generated using several techniques. We computed perturbation analysis and sequence analysis scores for each combination of prediction model and explanation algorithm.

*Table 4. Parameter values for explanation methods. KernelSHAP does not have any parameter to fix. The only parameter of TimeSHAP is the tolerance which is related to its pruning mechanism. Our algorithms parameters are described under section 3.3*

| Explanation Algorithm | Data sets | | |
| --- | --- | --- | --- |
| | TBI EHR data | TBI physiologic data | MIMIC-III |
| TimeSHAP | $Tolerance = 0.05$ | $Tolerance = 0.05$ | $Tolerance = 0.05$ |
| Stationary WindowSHAP | $l = 20$ | $l = 14$ | $l = 5$ |
| Sliding WindowSHAP | $l = 15$ $s = 8$ | $l = 15$ $s = 8$ | $l = 10$ $s = 6$ |
| Dynamic WindowSHAP | $\delta = 0.01$ $n_w = 20$ | $\delta = 0.001$ $n_w = 14$ | $\delta = 0.001$ $n_w = 20$ |

\* $l$: Window length, $s$: Stride, $\delta$: Shapley value threshold, $n_w$: maximum number of windows

## 4 Results

In this section, we present the outcomes of quantitative analysis, computational complexity analyses, and qualitative analysis. As part of the quantitative analysis, we demonstrate how evaluation measures of *WindowSHAP* explanations compare against baseline approaches. Next, the computational complexity analysis demonstrates how *WindowSHAP* affects the runtime and memory utilization of Shapley value extraction. Finally, a qualitative comparison between KernelSHAP and WindowSHAP is offered based on the explanations of an exemplar patient record from the MIMIC-III dataset.

**Quantitative analysis:** As the output of all prediction models is binary, the binary cross-entropy loss function was utilized as the quality metric in calculating perturbation and sequence analysis scores. It is worth noting that as the loss function values rise, it indicates that the prediction model performs worse, and as a result, some downstream performance indicators such as accuracy fall. Therefore, the greater the percent change in loss function, the higher the explanation quality. **Figure 5** depicts the outcomes of experimental tests for three different datasets. *WindowSHAP* outperformed or provided similar results as compared to competitors (i.e., KernelSHAP and TimeSHAP).

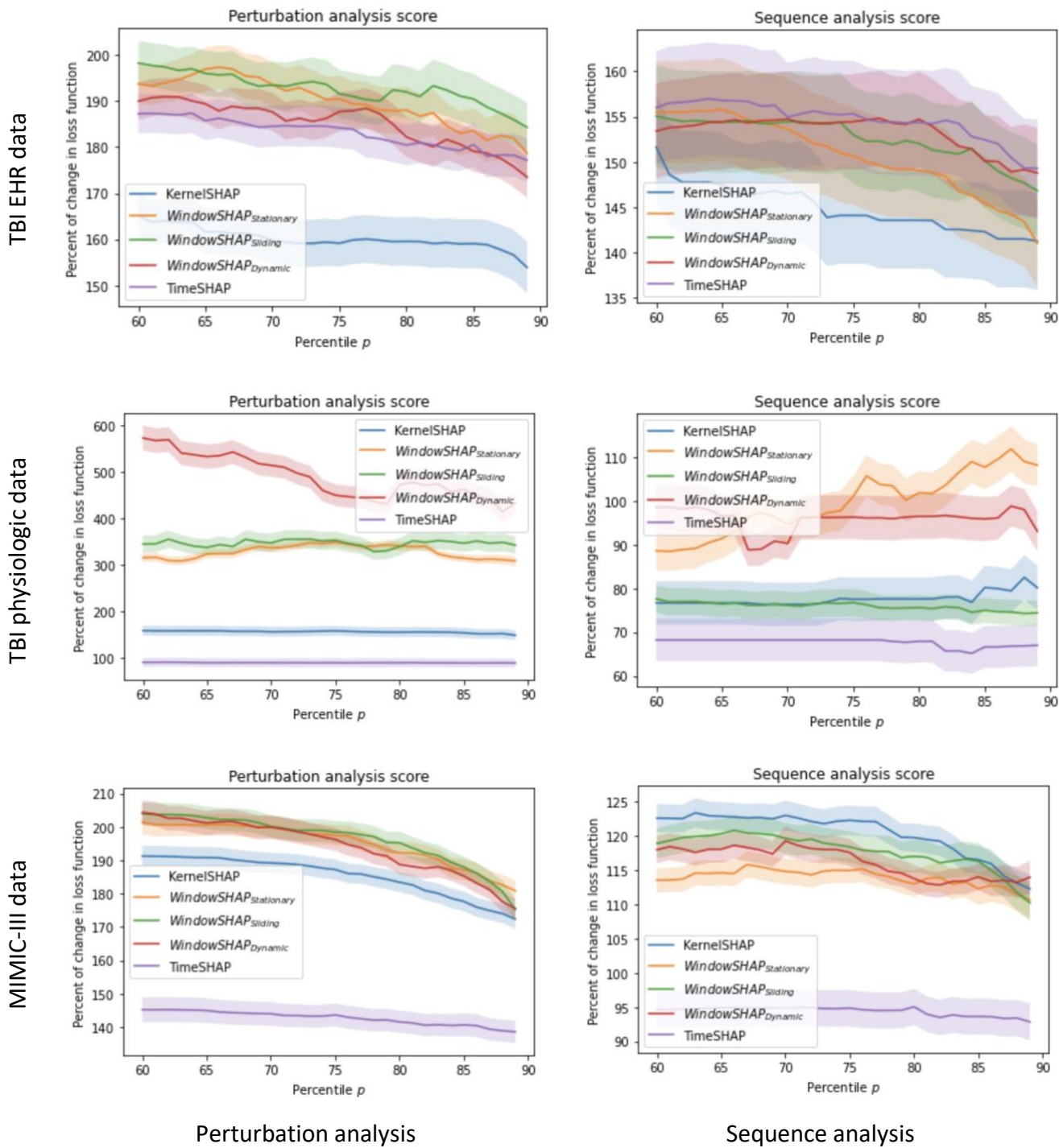

Perturbation analysis                                      Sequence analysis

*Figure 5.* Evaluation metrics for all explanation algorithms. Each row of figures shows the result for one of the prediction models. The x axis in all figures represents the percentile p that is used in the metrics definitions. The y axis represents the change in the quality metric after perturbing the most crucial time points. Error bars are shown as the mean ± standard errors of the mean of binary loss function.

**Computational complexity analysis:** The order of complexity for the WindowSHAP variants is less than KernelSHAP (**Table 5**).

Table 5. *Order of complexity of designed algorithms and the original implementation of Shapley values*

| KernelSHAP | Stationary WindowSHAP | Sliding WindowSHAP | Dynamic WindowSHAP |
|---|---|---|---|
| $O(D \times L2^{D \times L})$ | $O\left(D \left\lceil \frac{L}{l} \right\rceil 2^{D\left\lceil \frac{L}{l} \right\rceil - 1}\right)$ | $O\left(2D \left\lceil \frac{L-1}{s} \right\rceil 2^{2D-1}\right)$ | $O(D \cdot n_w^2 2^{D \cdot n_w - 1})$ |

We evaluated the memory usage and CPU time of each suggested algorithm for various hyperparameter values. **Figure 6** depicts the results of the complexity study performed on the TBI EHR data prediction model. The total number of variables is 62, and the length of time-series is 120 (each time step represents an hour). This data set was chosen to evaluate the computational complexity since it has the highest number of variable-time point combinations. It is noteworthy that the original implementation of Shapley values, KernelSHAP, has the same complexity as the *Stationary WindowSHAP* algorithm when the window length is one. In terms of computational complexity, *WindowSHAP* has significantly lower computational cost for generating Shapley values (**Figure 6**).

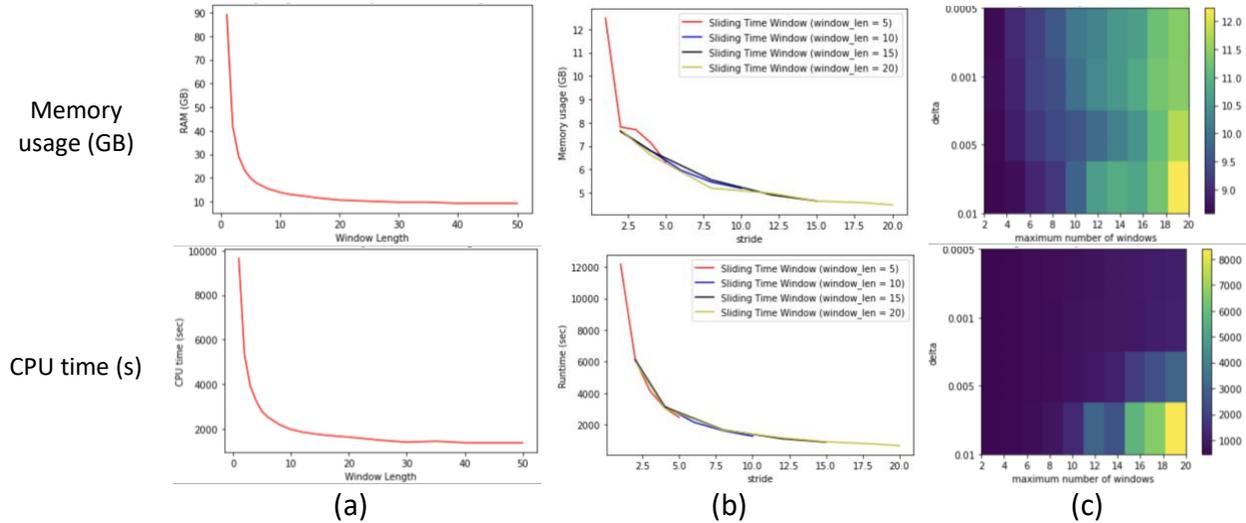

**Figure 6.** *Visualization of RAM usage and CPU time of different algorithms under WindowSHAP framework. Columns (a), (b), and (c) represent Stationary, Sliding, and Dynamic WindowSHAP algorithms respectively.*

**Qualitative analysis:** We use local explanations (i.e., the most important features) for a single patient record (see **Figure 7**) from the MIMIC-III dataset in order to illustrate how *WindowSHAP* differs from the original implementation of Shapley values, KernelSHAP. Based on the importance of each time step and variable, it is evident that the explanations of the two techniques are different. Dynamic WindowSHAP focuses more on the final time steps, whereas KernelSHAP assigns Shapley values to all time steps. **Figure 8** displays the findings of the two techniques' explanations for only heart rate variable.

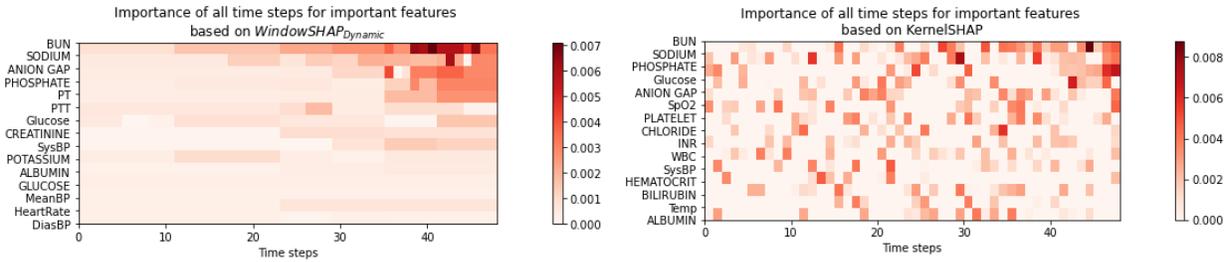

*Figure 7. Heatmaps depicting the importance of all time steps for the important features for a certain patient record from the MIMIC-III dataset. The top 15 variables depicted on the y axis are ranked according to their importance. The darker the color is, the higher the absolute value of the assigned Shapley value is.*

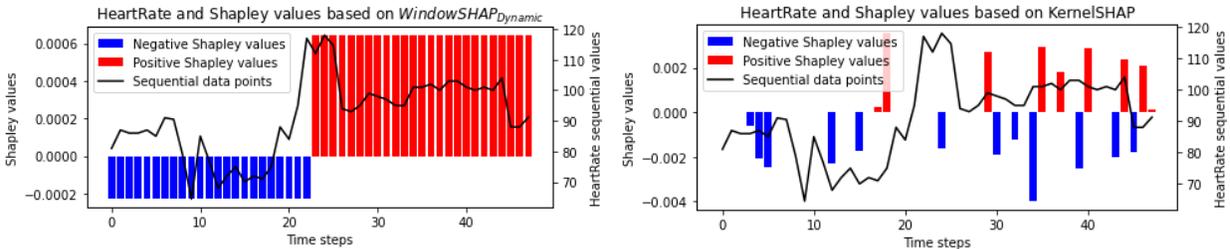

*Figure 8. The explanations of the heart rate variable for a patient in MIMIC mortality prediction model. The left and right plots represent visual explanations of WindowSHAP and KernelSHAP, respectively.*

## 5   Discussion

In order to interpret the outcomes of perturbation and sequence analysis metrics, it is crucial to acknowledge that these scores reflect the alteration in the prediction model's loss function when values at important time points are replaced with non-informative values. Consequently, explanation methods resulting in a more significant change in the loss function value are deemed superior, as they are likely to have more accurately identified the essential time steps and intervals based on the prediction model's behavior. As illustrated in **Figure 5**, *WindowSHAP* variants demonstrate superior explanatory performance for the majority of datasets and metrics by inducing a larger change in the loss function value upon perturbation of the most critical time steps and intervals. However, for the TBI EHR dataset, the performance of WindowSHAP variants and TimeSHAP is not significantly different, but both outperform KernelSHAP. Solely for the MIMIC dataset, based on the sequence analysis metric, KernelSHAP provides the highest quality explanation, albeit with a negligible difference. Nonetheless, employing WindowSHAP remains justifiable, as it delivers explanations considerably faster than KernelSHAP for lengthy time-series data.

TimeSHAP is developed under the premise that the initial time steps in time-series data are of lesser importance. Consequently, it aggregates the initial time steps and assigns them a single Shapley value. This could be the primary reason behind TimeSHAP's inferior performance for the TBI physiologic and MIMIC-III datasets. In contrast, *WindowSHAP* operates without such an assumption. For example, *Dynamic WindowSHAP* aggregates adjacent non-important time steps regardless of their position within the sequence, be it at the beginning or the end. This distinction may contribute to the differences in performance between the two methods.

The strong performance of *WindowSHAP* in terms of explainability can also be attributed to two factors. First, by aggregating nearby time steps as a time window, *WindowSHAP* lowers the dependence of the elements (i.e., time windows) for which the Shapley values are calculated, hence improving the performance of the explanation. Second, as illustrated in **Figure 1**, by aggregating neighboring time steps as time windows, the Shapley values of adjacent time steps might cancel each other out because their absolute values are nearly identical but in opposite directions. This results in an extremely low Shapley value in the associated time window, demonstrating its true insignificance.

For the TBI EHR data prediction model with 62 variables and 120 time steps, KernelSHAP requires approximately 90GB of RAM and $10^4$ seconds to calculate the Shapely values. By decreasing the duration of the time window in *Stationary WindowSHAP*, RAM and CPU time are reduced exponentially. For example, we show that merging 10 adjacent time points can reduce the CPU time by 80%. The complexity of *Stationary WindowSHAP* is dependent on the length of the time window, but the complexity of *Sliding WindowSHAP* is independent of length and only depends on the stride value. However, both $\delta$ and $n_w$ parameters affect the complexity of *Dynamic WindowSHAP*.

The explanations of KernelSHAP and *WindowSHAP* are different from each other. The results for explanation related to mortality prediction from the MIMIC-III dataset demonstrate that while KernelSHAP assigns large Shapley values to nearly all of the feature space and the complete time spectrum, *Dynamic WindowSHAP* assigns greater Shapley values to the final time steps, as this is more realistic and logical. Further, in *Dynamic WindowSHAP*, the length of time windows increases as the variables become less significant (lower on the $y$ axis in **Figure 7),** hence avoiding the calculation of Shapley values for less significant time points. The calculated Shapley values for heart rate shows that *Dynamic WindowSHAP* separates the sequence into two sections and only calculates two Shapley values, whereas KernelSHAP assigns single positive and negative values to some of the time points. Based on *Dynamic WindowSHAP*, the early part of the sequence contributes more to the unfavorable outcome (survival), whereas the second half contributes more to the positive outcome (death). Since the second half has a higher heart rate than the first, it is more rational to assign two opposite Shapley values to each segment. This is an illustration of how assigning Shapley values to windows, as opposed to scattered time points, makes explanations more comprehensible for end-users. The reason the algorithm only allocates two windows to this sequence is because the overall contribution of each window is not significant enough (less than $\delta$) so the algorithm does not further split them.

*WindowSHAP* has significant clinical implications, as it can assist users in better comprehending complex time-series data obtained from electronic health records (EHRs), physiological monitoring devices, and other sources. By identifying critical time points and temporal patterns using the *WindowSHAP* framework, clinicians can gain valuable insights into underlying clinical processes and relationships. For example, *WindowSHAP* can help uncover hidden temporal patterns in physiological data, such as vital signs or lab results, which may be indicative of disease progression or response to treatment. Additionally, it can be used to identify crucial time intervals in EHR data, shedding light on the relationship between specific medical events and patient outcomes. This, in turn, can guide clinicians in making more informed decisions about treatment strategies or intervention timing. In addition, the enhanced interpretability provided by *WindowSHAP* can help bridge the gap between sophisticated machine learning models and clinical decision-making, fostering clinicians' trust and confidence in machine learning based tools.

*Limitations:* The need to tune the parameters of each algorithm is one of the limitations of this study. We utilized a basic grid search to determine the optimal explanation algorithm parameters. However, outcomes may vary from one data set to another or even within data instances. As an example, since *Dynamic WindowSHAP* adheres to the local accuracy property, the prediction outcome would equal the sum of all Shapley values. Consequently, the effect of the threshold value ($\delta$) on the quality of the explanation is dependent on the model outcome. In other words, even for the same prediction model, a single $\delta$ would not provide decent explanations for different data instances. One of the potential future developments for the *Dynamic WindowSHAP* method is to dynamically calculate the threshold based on the model output. The fact that *WindowSHAP* algorithms are designed to behave similarly for different variables is one of its constraints. For instance, in the *Stationary WindowSHAP* technique, the length of time windows is the same for all variables in the time-series data, despite the fact that it may seem necessary to have varied window lengths for variables based on their relevance or rate of change.

## 6 Conclusion

Clinical machine learning models have strong prediction accuracies, but they are opaque because of their non-linear hierarchical structure, making it difficult to determine what details in the input data are causing specific predictions. While considerable effort has gone into understanding deep learning models, time-series models have received comparatively little attention. Our *WindowSHAP* framework offers a promising way for understanding the behavior of all forms of time-series classifiers. Three distinct algorithms are created within the *WindowSHAP* framework and compared against baselines. The results demonstrate that the explanations provided by our algorithms are of greater quality, i.e., by perturbing the most important time points based on our explanations, the performance of prediction models decreases more. For instance, for the TBI physiologic data, our algorithms' explanations identify the most significant time points that, if perturbed, would result in a rise in the loss function that is greater than twice that of baseline techniques. Our study also demonstrates that by utilizing the *WindowSHAP* framework, the computational complexity related to explainability can be dramatically reduced.

### Acknowledgments

This material is based upon work supported by the National Science Foundation under grants #1838730 and #1838745. Dr. Foreman was supported by the National Institute of Neurological

Disorders and Stroke of the National Institutes of Health (K23NS101123). The content is solely the responsibility of the authors. Any opinions, findings, and conclusions or recommendations expressed in this material are those of the authors and do not necessarily reflect the views of the National Science Foundation or of the National Institutes of Health. The authors acknowledge the TRACK-TBI Study Investigators for providing access to data used in this work.